\documentclass[runningheads]{llncs}

\usepackage{eccv}

\usepackage{eccvabbrv}

\usepackage{graphicx}
\usepackage{booktabs}

\usepackage[accsupp]{axessibility}  %

\usepackage{hyperref}

\usepackage{orcidlink}
\usepackage{caption}
\usepackage{graphicx}
\usepackage{wrapfig}
\usepackage{makecell}
\usepackage{microtype}
\usepackage{adjustbox}
\usepackage{multirow}
\usepackage{colortbl}
\usepackage{multicol}
\usepackage{kotex}
\usepackage{pifont}

\begin{document}

\title{Rethinking Detection Calibration:\\A Coordinate and Direction Perspective} 

\author{Juyong Lee\inst{1}\orcidlink{0009-0006-9054-7756} \and
Seungjin Jung\inst{1}\orcidlink{0009-0009-7853-2588} \and \\
Jungmin Lee\inst{2}\orcidlink{0009-0009-1172-7209} \and 
Sunju Lee\inst{1}\orcidlink{0009-0006-6641-7038} \and
Jongwon Choi \inst{1,2,3}\orcidlink{0000-0001-9753-8760}\thanks{Corresponding author}}

\authorrunning{Lee et al.}

\institute{Dept. of AI, Chung-Ang University, Republic of Korea \and
Dept. of Advanced Imaging, GSAIM, Chung-Ang University, Republic of Korea \and
GS. of Virtual Convergence,
Chung-Ang University, Republic of Korea \\
\email{\{jylee, sjjung, jngmnlee, lsjoo\}@vilab.cau.ac.kr, choijw@cau.ac.kr}}

\maketitle

\begin{abstract}
Deep learning based object detectors require trustworthiness beyond competitive detection performance, but deep neural networks are prone to overconfident predictions, assigning high confidence scores to predictions that are likely to be inaccurate. To improve the alignment between confidence scores and prediction accuracy, existing methods calibrate confidence scores based on box-level localization, such as precision or intersection over union with the ground truth bounding box. However, box-level localization reflects only a measure of agreement between the predicted box and the ground truth, resulting in calibrated confidence scores for box-level accuracy failing to capture the localization accuracy of coordinates of box. To tackle this issue, we propose a novel post-hoc calibration framework, rethinking detection calibration (ReDC), which provides reliable coordinate-level confidence scores, including directional information. The proposed framework defines coordinate-wise alignment and deviation direction between predictions and ground truth. Based on the alignment measure, confidence re-encoding produces reliable coordinate-level confidence scores, while directional displacement estimation predicts coordinate-wise deviation directions. Extensive experiments under in-domain and out-domain scenarios demonstrate that the proposed approach expresses the coordinate-wise localization of detected objects more precisely than existing methods. Furthermore, our method covers the representational scope of prior calibration approaches by aggregating coordinate-level confidence scores into box-level localization.
\keywords{Confidence calibration \and Object detection \and Explainable computer vision}
\end{abstract}

\section{Introduction}\label{sec:intro}

\begin{figure}[t]
     \centering
     \includegraphics[width=1\textwidth]{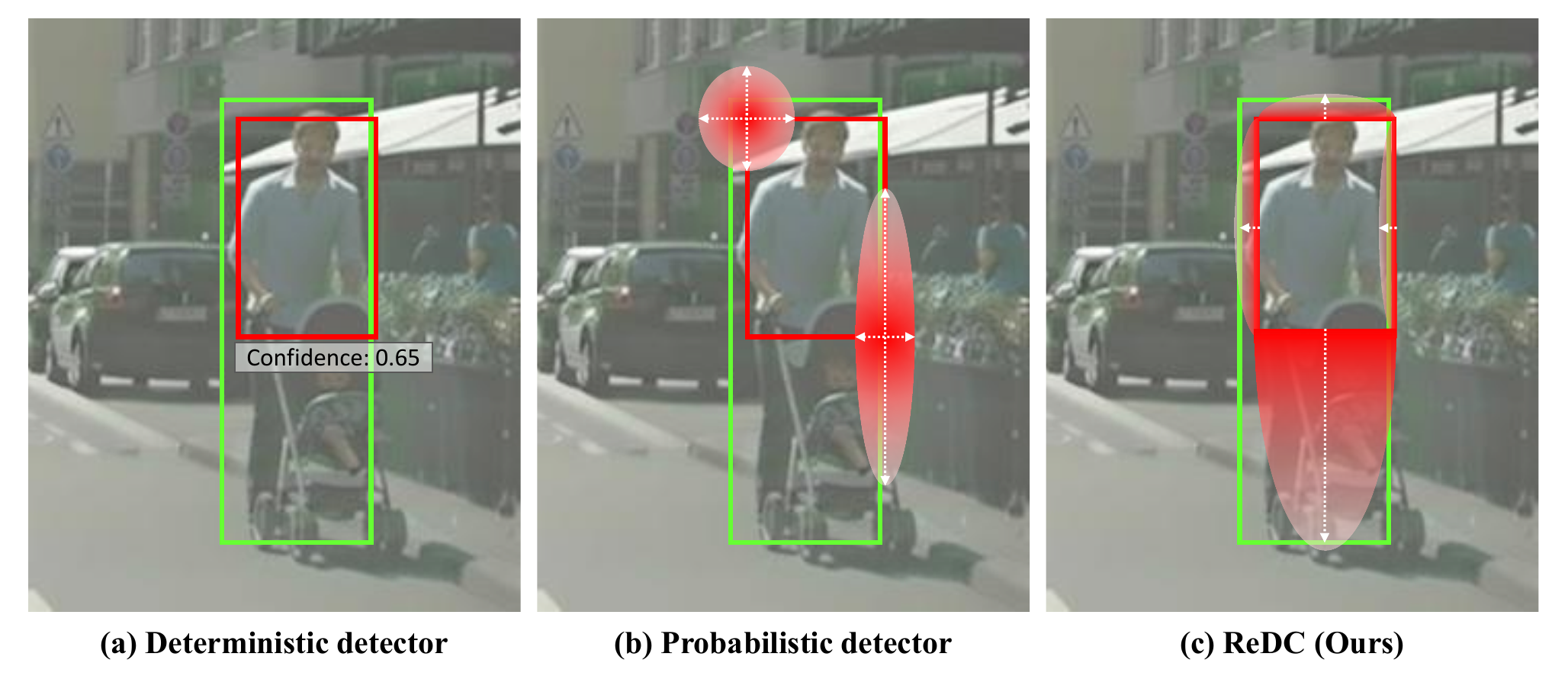}
     \caption{
        \textbf{Conceptual comparison of calibration frameworks in object detection.} (a) Deterministic detector: Conventional methods provide a single confidence score for the entire bounding box, failing to identify specific coordinate-wise misalignments. (b) Probabilistic detector: While capturing local confidence, symmetric distributions often fail to represent the specific direction and magnitude of coordinate deviation. (c) Rethinking Detection Calibration (ReDC): Our framework estimates independent coordinate-wise confidence scores aligned with the coordinate-wise alignment ratio (CAR). This approach explicitly captures directional confidence.}
     \label{fig:Teaser}
 \end{figure}
 
Object detection has demonstrated impressive performance through progressive advances~\cite{zhu2020deformable, zhang2023dino}. However, deep neural networks are often overconfident~\cite{guo2017calibration, zhang2020mix}, which may result in incorrect predictions made with high confidence and limit the practical adoption of object detection in safety-critical domains such as autonomous driving~\cite{wu2017squeezedet,feng2021review}, medical applications~\cite{zhang2020generalizing,jaeger2020retina}, and security systems~\cite{liu2018learning}. To assign lower confidence scores to predictions that are likely to be inaccurate, confidence calibration~\cite{guo2017calibration,tomani2022parameterized,pmlr-v202-jung23a} has been widely studied to align predicted confidence scores with their empirical accuracies. In object detection, confidence calibration further requires properly defining empirical accuracy for bounding box localization~\cite{munir2023bridging,kuzucu2024calibration, munir2023caldetr}, as deterministic detectors do not provide probabilistic estimates for bounding boxes.

Previous studies typically define empirical accuracy at the box-level. One approach measures box-level accuracy using precision with a fixed intersection over union (IoU) threshold~\cite{oksuz2023towards}, while another incorporates the predicted IoU value to estimate box-level accuracy~\cite{kuzucu2024calibration}. However, object detection predictions can exhibit different degrees of spatial misalignment and directional shifts relative to the ground-truth bounding box, even when they share identical box-level accuracy. As illustrated in Fig.~\ref{fig:Teaser}-(a), existing confidence calibration methods capture the degree of alignment only at the bounding-box level because they rely solely on box-level localization accuracy, limiting their ability to reflect fine-grained coordinate-wise misalignment. This limitation can also affect downstream decision-making in real-world applications.

Probabilistic object detection methods~\cite{Kueppers_2022_ECCV_Workshops} estimate coordinate-wise covariance to derive confidence, defining empirical accuracy as whether the ground-truth box falls within the predicted probabilistic region (Fig.~\ref{fig:Teaser}-(b)). This captures coordinate-wise localization error but, since coordinate-wise estimates are collapsed into box-level confidence and accuracy, still fails to capture the direction of localization error. This matters in practice: in autonomous driving, a box shifted upward and smaller than ground truth can make an object appear farther away, potentially delaying braking, despite identical box-level accuracy. Moreover, these methods require a probabilistic detector, limiting general applicability.

To address the limitations in prior approaches, we propose a new post-hoc calibration framework without restricting its applicability to specific object detectors, rethinking detection calibration (ReDC), consisting of confidence re-encoding (CR), coordinate-wise alignment ratio (CAR), and directional displacement estimation (DDE). CR estimates coordinate-wise confidence by re-scaling the class confidence for each coordinate, while CAR computes coordinate-wise empirical accuracy using the bounding boxes of the ground truth and the instance sample. Furthermore, DDE estimates the deviation direction of each predicted coordinate by utilizing predicted logit vectors and information derived from coordinates. By aligning the resulting confidence and empirical accuracy, the model learns to capture spatial misalignment and directional shifts as illustrated in Fig.~\ref{fig:Teaser}-(c).
Since coordinate-wise alignment alone cannot capture box-level accuracy, we approximate box-level IoU from coordinate-wise confidence and deviation direction, and train a calibration network to align this estimated IoU with actual IoU. This gives ReDC complementary calibration at both the coordinate and box levels. We also introduce two new metrics, as no prior metric jointly captures coordinate-level and directional accuracy: coordinate-wise expected calibration error (C-ECE) for coordinate-level calibration quality, and direction-aware calibration error (Da-CE) for calibration performance incorporating deviation direction.

We compare ReDC against state-of-the-art post-hoc and train-time calibration methods on COCO~\cite{lin2014microsoft} and Cityscapes~\cite{Cordts2016Cityscapes} under both in-domain and out-of-domain settings. Results show ReDC effectively captures coordinate-wise localization accuracy and deviation direction while maintaining competitive box-level calibration performance.

The main contributions of this paper are summarized as follows:
\begin{itemize}
    \item We analyze the heterogeneity of coordinate-level accuracy among samples sharing similar box-level accuracy and show that existing calibration methods exhibit limitations in independently representing coordinate-level accuracy.
    
    \item We propose ReDC, a post-hoc calibration framework that conveys coordinate-wise confidence scores in object detection while incorporating directional information of prediction misalignment with respect to the ground truth.
    
    \item Extensive experiments demonstrate that ReDC outperforms existing calibration methods in fine-grained calibration and achieves comparable box-level calibration performance by aggregating fine-grained confidence scores to approximate IoU.
\end{itemize}

\section{Related Work}\label{sec:Relwork}
\subsection{Confidence Calibration}
Calibration methods aim to align model confidence with empirical accuracy. Temperature scaling (TS)~\cite{guo2017calibration}, a simple yet effective post-hoc method, has been widely adopted in subsequent works~\cite{pmlr-v202-jung23a,zhang2020mix,tomani2022parameterized}. In object detection, confidence calibration additionally requires properly defining empirical accuracy for bounding box localization~\cite{pathiraja2023multiclass,munir2023bridging}, as deterministic detectors do not naturally provide probabilistic estimates for bounding boxes. Existing approaches address this problem through train-time calibration objectives integrated into detector training or by aligning detection confidence with localization quality~\cite{munir2023caldetr}. Post-hoc calibration methods have also been explored, including covariance calibration for bounding box distributions~\cite{Kueppers_2020_CVPR_Workshops}, confidence calibration for object detectors~\cite{pan2021model}, and calibration analysis with evaluation baselines for object detection~\cite{oksuz2023towards,kuzucu2024calibration}. However, these approaches primarily operate at the box level, calibrating detection confidence without modeling coordinate-level localization information. In contrast, our method calibrates the coordinate-level confidence score and leverages it to estimate the box-level localization score.

\subsection{Coordinate-wise Uncertainty Estimation}
Recent studies estimate localization uncertainty in object detection by modeling bounding box coordinates as probabilistic distributions. Several works predict Gaussian distributions over bounding box coordinates to estimate coordinate-wise uncertainty~\cite{klloss,Choi_2019_ICCV,harakeh2020bayesod}, while other approaches extend uncertainty estimation to anchor-free detectors~\cite{lee2022uad}. More recent studies focus on uncertainty calibration to align predicted uncertainty with empirical errors~\cite{song2019distribution,Kueppers_2022_ECCV_Workshops}. However, existing approaches typically convert localization uncertainty into box-level accuracy and confidence for calibration, limiting their ability to capture directional localization errors. Moreover, since modeling coordinate-wise uncertainty as a probabilistic distribution requires probabilistic detectors, its applicability to commonly used deterministic detectors is limited. In contrast, our method first calibrates coordinate-level confidence scores and then estimates box-level confidence score. Moreover, we learn a lightweight post-hoc module without retraining the detector or relying on specific detectors, improving training efficiency.

\section{Preliminary}\label{sec:pre}

\subsection{Confidence Calibration for Object Detection}
\subsubsection{Object Detection}
We define a dataset $\mathcal{D} := \{(\mathbf{x}_i, \mathbf{y}_i)\}_{i=1}^N$, where $i$ denotes the sample index and $N$ denotes the total number of samples. Here, $\mathbf{x}_i \in \mathbb{R}^{H \times W \times Z}$ denotes an input image with height $H$, width $W$, and $Z$ color channels, while the corresponding label $\mathbf{y}_i = (c_i, b_i)$ consists of a class label $c_i$ and a bounding box $b_i \in \mathbb{R}^4$. Given an input image $\mathbf{x}_i$, an object detector $\phi_D$ produces $M$ detections, defined as follows:

\begin{equation}
    \phi_D(\mathbf{x}_i) : \mathbf{x}_i \mapsto \{(\hat{c}_j, \hat{b}_j, \hat{p}_j)\}_{j=1}^M,
\end{equation}
where $\hat{c}_j$, $\hat{b}_j$, and $\hat{p}_j$ denote the predicted class label, bounding box, and predicted confidence score for the object, respectively.

\subsubsection{Post-hoc Calibration}
Post-hoc confidence calibration aligns prediction confidence scores $\hat{p}$ with the true probabilities $p$ by training an additional calibration model added to a pretrained model.
In classification tasks, perfect calibration~\cite{guo2017calibration} is defined as follows:
\begin{equation}
    \mathbb{P}(\hat{y}=y\mid\hat{p}=p)=p, \quad\forall p\in[0,1],
\end{equation}
where $\hat{y}$ denotes predicted class.
Since the true conditional probability is not directly observable, calibration is evaluated in practice using empirical accuracy estimated from samples with similar confidence scores.
In object detection, Kuppers et al.~\cite{Kueppers_2020_CVPR_Workshops} proposed training calibrators for the object detection task to satisfy the following equation by setting precision as the accuracy:
\begin{equation}
    \mathbb{P}(m = 1 \mid \hat{c} = c, \hat{b} = b, \hat{p} = p) = p, \quad \forall p \in [0,1],
\end{equation}
where $\hat{p}$ denotes the confidence score associated with a predicted bounding box inferred by the object detector, and $m \in \{0,1\}$ indicates whether the detector correctly identifies a ground-truth object.
In practice, $m$ is determined based on an intersection over union (IoU) criterion with a threshold $\tau$.
To provide information on localization accuracy, Kuzuku et al.~\cite{kuzucu2024calibration} proposed setting the IoU threshold $\tau$ to 0 and aligning confidence scores with IoU, which yields the following equation: 
\begin{equation} 
\mathbb{E}_{\hat{b}\in\ B(\hat{p})}[\text{IoU}(\hat{b},b)]=\hat{p},\quad \forall\hat{p}\in[0,1], 
\end{equation} 
where $B(\hat{p})$ denotes the set of predicted bounding boxes with confidence score $\hat{p}$, and $b$ denotes the ground-truth bounding box matched to $\hat{b}$.
\begin{figure}[t]
     \centering
     \includegraphics[width=1\linewidth]{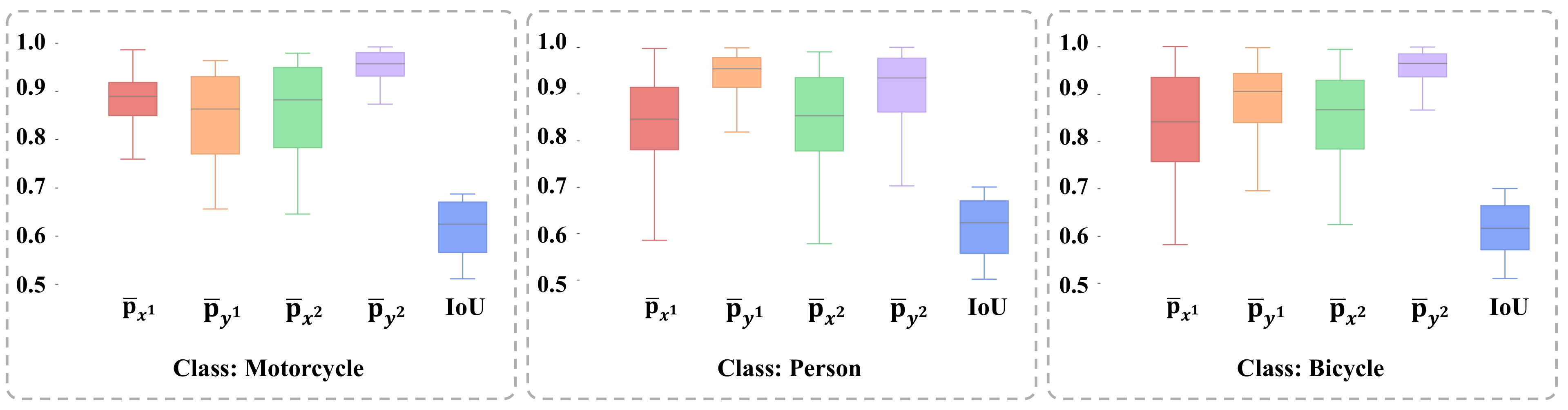}
     \caption{
        \textbf{Box-level IoU fails to capture coordinate-wise alignments.} From left to right, the figures correspond to the motorcycle, person, and bicycle classes, respectively. Each box plot shows the distribution of coordinate-wise alignment ratios (CAR) computed over samples belonging to the same class, where CAR measures the degree of correspondence between predicted and ground-truth bounding-box coordinates. 
    }
    \label{fig:Prelim}
\end{figure}
\begin{figure}[t!]
    \centering
    \includegraphics[width=1\textwidth]{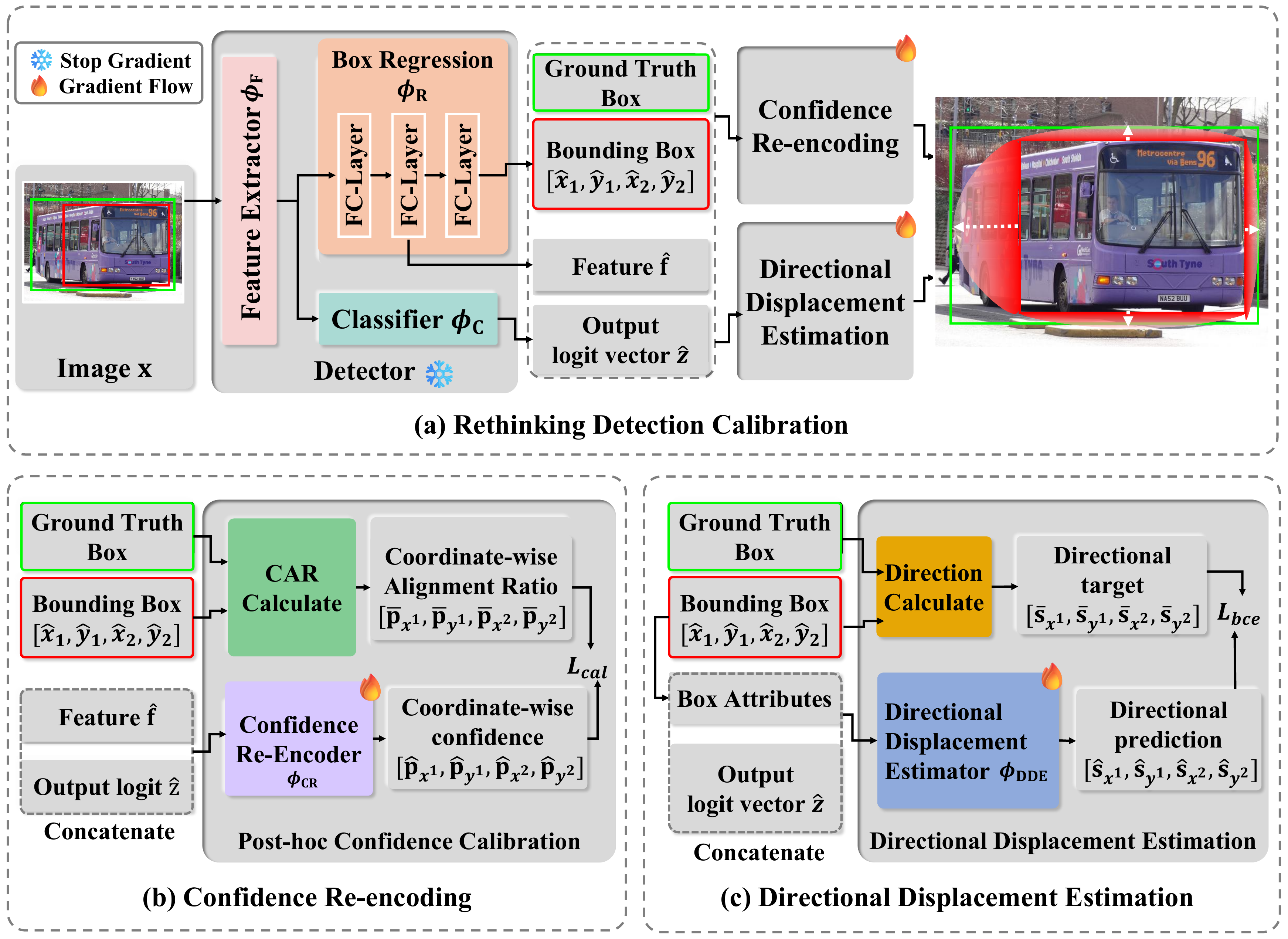}
    \caption{
        \textbf{Overview of ReDC framework.} (a) Rethinking detection calibration is a post-hoc calibration method that estimates coordinate-wise confidence scores and deviation directions on top of a pre-trained detector. (b) Confidence re-encoding estimates coordinate-wise confidence scores aligned with coordinate-wise alignment ratio, which represents coordinate-wise accuracy. (c) Directional displacement estimation predicts the direction in which each predicted coordinate deviates from the corresponding ground-truth coordinate.
    }
    \label{fig:frame}
\end{figure}

\subsection{Preliminary Analysis\label{sec:PreliminaryAnalysis}}
This section describes the limitation that existing object detection calibration methods do not sufficiently explain the localization accuracy of detection results. We analyze the localization alignment characteristics of bounding boxes at the coordinate level.

As shown in the Fig.~\ref{fig:Prelim}, the coordinate-wise alignment distributions are not uniform within the same class. Some coordinates exhibit higher variance and lower alignment, indicating that localization accuracy differs across bounding box coordinates. This behavior is more evident for objects with complex geometries. Such coordinate-specific errors are more pronounced when objects have complex structures or are affected by viewpoint changes.

However, existing confidence calibration methods for the object detection task treat the entire bounding box as a single unit and learn a single confidence score to represent localization accuracy. This approach fails to capture coordinate-wise variations in alignment and does not adequately reflect severe localization errors occurring in specific coordinates. These observations suggest the need for coordinate-wise localization confidence estimation.

\section{Method}\label{sec:Met}
\subsection{Rethinking Detection Calibration}

Motivated by the analysis in Sec.~\ref{sec:PreliminaryAnalysis}, we introduce a new post-hoc calibration framework, termed rethinking detection calibration (ReDC). 
ReDC consists of an confidence re-encoder (CR) that re-encodes detection information to calibrate confidence scores for coordinate-wise localization and a directional displacement estimator (DDE) that estimates the direction of coordinate-wise deviation, as illustrated in Fig.~\ref{fig:frame}. 

\begin{wrapfigure}{t!}{0.48\linewidth}
    \centering
    \includegraphics[width=\linewidth]{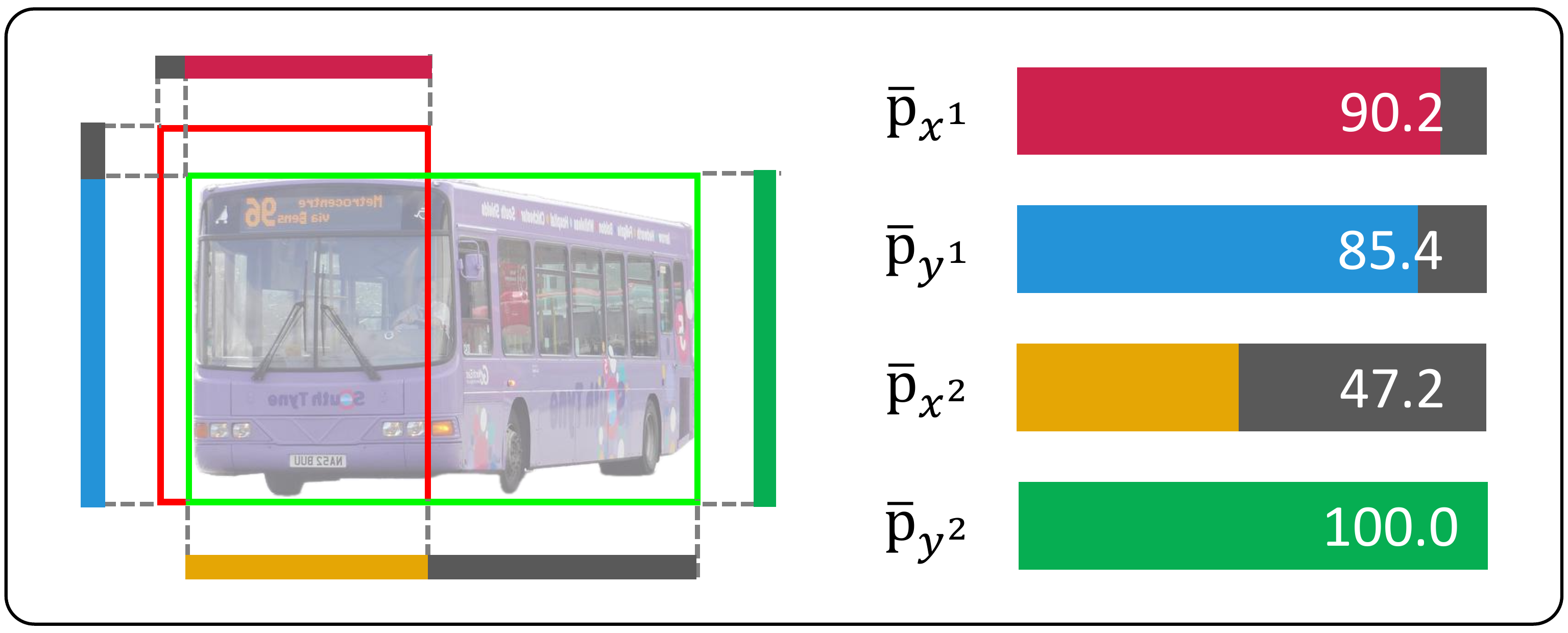}
    \caption{\textbf{Coordinate-wise alignment ratio (CAR).} CAR measures the coordinate-wise alignment between \textcolor{red}{predicted} and \textcolor{green}{ground-truth} bounding boxes based on the coordinate-wise absolute difference and axis-wise intersection length.}\label{fig:car}
\end{wrapfigure}

\subsubsection{Coordinate-wise Alignment Ratio}
\label{subsub:cdiou}
We propose a coordinate-wise alignment ratio (CAR) to define the localization accuracy for each coordinate. CAR is computed based on coordinate-wise differences and intersection lengths between the predicted and ground-truth bounding boxes, as illustrated in Fig.~\ref{fig:car}. First, we consider the four bounding box coordinates corresponding to the top-left and bottom-right corners, \textit{i.e.}, $(x^{1}, y^{1})$ and $(x^{2}, y^{2})$, where $x^{2} = x^{1} + w$ and $y^{2} = y^{1} + h$, respectively. The predicted coordinates are obtained via a feature extractor $\phi_F$ and a box regression module $\phi_R$ of the pretrained object detector $\phi_D$, and are defined as $(\hat{x}^1, \hat{y}^1, \hat{w}, \hat{h}) = \phi_R(\phi_F(\mathrm{x}))$. Based on the predicted and ground-truth coordinates, the coordinate-wise differences are formulated as follows:
\begin{equation}
    \mathrm{dist}_{x^l} = |\hat{x}^l - x^l|, \quad
    \mathrm{dist}_{y^l} = |\hat{y}^l - y^l|,
\end{equation}
where $l \in \{1,2\}$ indexes the top-left ($l=1$) and bottom-right ($l=2$) coordinates of a bounding box. 
The $x$- and $y$-axis intersection lengths are defined as follows:
\begin{equation}
\begin{aligned}
&\mathrm{inter}_w =
\max\!\left(0, \min(\hat{x}^2, x^2) - \max(\hat{x}^1, x^1)\right), \\
&\mathrm{inter}_h =
\max\!\left(0, \min(\hat{y}^2, y^2) - \max(\hat{y}^1, y^1)\right).
\end{aligned}
\label{eqn:inter_length}
\end{equation}

Using the coordinate-wise differences and intersection lengths defined above,
we compute the CAR  $\bar{\mathrm{p}}=(\bar{\mathrm{p}}_{x^1},\bar{\mathrm{p}}_{y^1},\bar{\mathrm{p}}_{x^2},\bar{\mathrm{p}}_{y^2})$ as follows:
\begin{equation}
    \bar{\mathrm{p}}_{x^l}:=\frac{\mathrm{inter}_w}{\mathrm{dist}_{x^l}+\mathrm{inter}_w}, \quad
    \bar{\mathrm{p}}_{y^l}:=\frac{\mathrm{inter}_h}{\mathrm{dist}_{y^l}+\mathrm{inter}_h}.
    \label{Eq:CAR}
\end{equation}

Each CAR value lies in the range $[0,1]$.
For example, $\bar{\mathrm{p}}_{x^1}\!=\!0$ when the intersection width $\mathrm{inter}_w$ is zero, indicating no horizontal overlap between the two bounding boxes.
In contrast, $\bar{\mathrm{p}}_{x^1}\!=\!1$ when the left $x$-coordinates of the predicted and ground-truth bounding boxes exactly match.
Based on CAR, we require the following definition to be satisfied at the coordinate-wise level, extending the box-level formulation of Kuzucu et al.~\cite{kuzucu2024calibration}. Given the coordinate-wise confidence scores $\hat{\mathrm{p}}=(\hat{\mathrm{p}}_{x^1},\hat{\mathrm{p}}_{y^1},\hat{\mathrm{p}}_{x^2},\hat{\mathrm{p}}_{y^2})$, the $\hat{\mathrm{p}}_{t}$ is perfectly calibrated if satisfying the following equation:
\begin{equation}
    \underset{\hat{b}\in\ B(\hat{\mathrm{p}}_{t})}{\mathbb{E}}[\bar{\mathrm{p}}_{t}]=\hat{\mathrm{p}}_{t}, \quad \forall \hat{\mathrm{p}}_{t}\in[0,1], t\in\{x^1.y^1,x^2,y^2\}
    \label{eqn:perfect_car}
\end{equation}
where $B(\hat{\mathrm{p}}_{t})$ denotes the set of predicted bounding boxes $\hat{b}$ whose coordinate-wise confidence scores are equal to $\hat{\mathrm{p}}_{t}$, and $\bar{\mathrm{p}}_{t}$ denotes the coordinate-wise true probability associated with $\hat{b}$ as defined in Eq.~\ref{Eq:CAR}.

\subsubsection{Confidence Re-encoding}
\label{subsub:UR}
We designed CR to estimate coordinate-wise confidence scores.
To estimate coordinate-wise confidence scores using CR, we compute the following two feature representations: the output logit and the bounding box feature representation.
The output logit of $i$-th prediction is defined as $\hat{\mathrm{z}}_i = \phi_C(\phi_F(\mathbf{x}_i))$. 
The bounding box feature representation is defined as $\hat{\mathbf{f}}_i = \phi_R{[:-2]}(\phi_F(\mathbf{x}_i))$. To calibrate confidence scores with CAR, we leverage the bounding box feature $\hat{\mathbf{f}}_i$ to encode localization confidence to logit $\hat{\mathrm{z}}_i$. 
We obtain coordinate-wise confidence scores through CR $\phi_{\mathrm{CR}}$ formulated as follows:
\begin{equation}
    \hat{\mathbf{p}}_i 
    = (\hat{\mathrm{p}}_{({x^1},i)}, \hat{\mathrm{p}}_{({y^1},i)}, \hat{\mathrm{p}}_{({x^2},i)}, \hat{\mathrm{p}}_{({y^2},i)})
    = \sigma(\frac{\hat{\mathrm{z}}_i}{\phi_{\mathrm{CR}_{t}}(\hat{\mathbf{f}}_i)}+\beta_t), \quad t\in\{x_1.y_1,x_2,y_2\}.
    \label{eq:coordi-conf}
\end{equation}

Finally, we train CR $\phi_{\mathrm{CR}}$ to satisfy Eq.~\ref{eqn:perfect_car} by aligning the coordinate-wise confidence scores (Eq.~\ref{eq:coordi-conf}) with the corresponding the empirical coordinate-wise accuracies defined by CAR (Eq.~\ref{Eq:CAR}).
The object loss is derived from the negative log-likelihood (NLL) and is formulated as follows:
\begin{equation}
    \mathcal{L}_{cal}:=\mathbb{E}\left[-(\mathrm{\bar{p}}\log(\hat{\mathrm{p}})+(1-\mathrm{\bar{p}})\log(1-\hat{\mathrm{p}}))\right].
\end{equation}

\subsubsection{Directional Displacement Estimation}
We propose a directional displacement estimation that estimates the relative directional displacement of each predicted bounding box coordinate with respect to the ground truth bounding box. The proposed method determines whether the ground truth coordinate is larger or smaller than the predicted coordinate and thereby identifies the direction of the coordinate error. The directional displacement estimator $\phi_{\mathrm{DDE}}$ uses an MLP that takes the predicted logit vector $\hat{\mathbf{z}}_i$ and geometric attributes of the bounding box as input, including the center coordinates $\hat{\mathrm{cx}}_i$ and $\hat{\mathrm{cy}}_i$, width $\hat{\mathrm{w}}_i$, height $\hat{\mathrm{h}}_i$, area $\hat{\mathrm{A}}_i$, and width-to-height ratio $\hat{\mathrm{R}}_i$. The directional targets of each $t$ coordinate $\bar{\mathrm{s}}_t$ is defined as follows:

\begin{equation}
\bar{\mathrm{s}}_{t} =
\begin{cases}
+1 & \text{if } \hat{t} - t > 0 \\
-1 & \text{otherwise}
\end{cases}
 ,t\in\{x^1.y^1,x^2,y^2\}.
\end{equation}

In addition, the directional displacement estimator $\phi_{\mathrm{DDE}}$ produces the directional prediction $\hat{\mathbf{s}}_i$ as follows:
\begin{equation}
    \hat{\mathbf{s}}_i 
    = (\hat{\mathrm{s}}_{({x^1},i)}, \hat{\mathrm{s}}_{({y^1},i)}, \hat{\mathrm{s}}_{({x^2},i)}, \hat{\mathrm{s}}_{({y^2},i)})
    = \phi_{\mathrm{DDE}}(\hat{\mathbf{z}}_i, \hat{\mathrm{cx}}_i, \hat{\mathrm{cy}}_i, \hat{\mathrm{w}}_i, \hat{\mathrm{h}}_i, \hat{\mathrm{A}}_i, \hat{\mathrm{R}}_i).
    \label{eq:SE}
\end{equation}
The final directional prediction is defined as:
\begin{equation}
\hat{\mathrm{s}}_{(t,i)} =
\begin{cases}
+1 & \text{if } \sigma(\hat{\mathrm{z}}^c_{s_{(t,i)}}) \ge \tau_t^c \\
-1 & \text{otherwise}
\end{cases}
\quad ,t\in\{x^1,y^1,x^2,y^2\},
\end{equation}
where $\hat{\mathrm{z}}^c_{s_{(t,i)}}$ denotes the $i$-th predicted logit for class $c$ obtained through $\phi_{\mathrm{DDE}}$, $\sigma(\cdot)$ denotes the sigmoid function, and $\tau_t^c$ denotes the class-specific threshold for class $c$, determined based on the direction accuracy during the training process of $\phi_{\mathrm{DDE}}$. We first train the confidence re-encoder and then train the directional displacement estimator. Since each deviation direction has a binary label of $+1$ or $-1$, the directional displacement estimator is trained with the binary cross-entropy loss to estimate coordinate-wise deviation directions. During inference, $\sigma(\hat{\mathrm{z}}^c_{s_{(t,i)}})$ is compared with the threshold $\tau_t^c$ found during training to obtain the predicted direction $\hat{\mathrm{s}}_{(t,i)}$.

Furthermore, to provide both coordinate-level and box-level calibrated confidence scores, we show that intersection over union (IoU) can be derived from CAR and directional information between predicted and ground-truth bounding boxes. When coordinate-wise expected calibration error (C-ECE) reaches the minimum value, the calibrated confidence score equals CAR. Therefore, IoU approximation uses coordinate-wise calibrated confidence scores with directional information, and the approximated IoU score is further calibrated with ground-truth IoU using isotonic regression~\cite{zadrozny2002transforming} and Platt scaling~\cite{platt1999probabilistic}.
More details are provided in the supplementary material.

\subsection{Coordinate-wise Calibration Metric}

\subsubsection{Coordinate-wise Expected Calibration Error}

To measure coordinate-wise calibration performance for Eq.~\ref{eqn:perfect_car}, this work proposes the coordinate-wise expected calibration error (C-ECE), a new metric that quantitatively evaluates the alignment with CAR and coordinate-wise confidence score. The equation is given as follows:

\begin{equation}
    \text{C$_{t}$-ECE}=\frac{1}{C}\sum_{c=1}^{C}\sum_{j=1}^{J} \frac{\lvert B_j^c \rvert}{\lvert B^c \rvert} 
    \Big\lvert \bar{\mathrm{p}}_{t}(B_j^c) - \hat{\mathrm{p}}_{t}(B_j^c) \Big\vert \quad ,t\in\{x^1.y^1,x^2,y^2\},
    \label{eqn:c_ece}
\end{equation}
where $t$ denotes the coordinates that represent the top-left and bottom-right, and the estimation follows a class-wise scheme to prevent a specific class from dominating the error. Moreover, the evaluation partitions continuous confidence scores into $J$ bins that are equally spaced over the confidence range. Accordingly, $B_j$ indicates the $j$-th bin, and $\hat{\mathrm{p}}_{t}(B_j^c)$ denotes the average coordinate-wise confidence score of the prediction belonging to $B_j^c$, while $\bar{\mathrm{p}}_{t}(B_j^c)$ represents the average CAR of the predictions in the corresponding bin.

\subsubsection{Direction-aware Calibration Error}

We introduce direction-aware calibration error (Da-CE) to evaluate coordinate-wise calibration error that incorporates directional information. The Da-CE evaluates the degree of misalignment between predicted bounding box coordinates and ground truth bounding box coordinates while accounting for directional deviation. We first define coordinate-wise mismatch between predicted and ground truth coordinates as:
\begin{equation}
    \bar{\mathrm{U}}_{(t,i)}=\bar{\mathrm{s}}_{(t,i)} \cdot (1- \bar{\mathrm{p}}_{(t,i)})\quad,t\in\{x^1.y^1,x^2,y^2\},
\end{equation}
where $\bar{\mathrm{p}}_t$ denotes the CAR between ground truth bounding box and predicted bounding box and $\bar{\mathbf{s}}_t$ indicates the direction of CAR. Furthermore, predicted mismatch that incorporates the predicted direction is defined as:
\begin{equation}
    \hat{\mathrm{U}}_{(t,i)}=\hat{\mathrm{s}}_{(t,i)} \cdot (1- \hat{\mathrm{p}}_{(t,i)} )\quad,t\in\{x^1.y^1,x^2,y^2\}.
\end{equation}

With these definitions, Da-CE is defined as:

\begin{equation}
    \text{Da$_{t}$-CE}=\frac{1}{C}\sum_{c=1}^{C}\sum_{j=1}^{J} \frac{\lvert B_j^c \rvert}{\lvert B^c \rvert} 
    \Big( \frac{1}{\lvert B_{j_\text{TP}}^c \rvert}\sum_{i\in B_{j_\text{TP}}^c}\lvert\bar{\mathrm{U}}_{(t,i)} - \hat{\mathrm{U}}_{(t,i)}\rvert \Big),
    \label{eqn:dauce}
\end{equation}
where $t$ denotes a coordinate, i.e., $t \in {x^1, y^1, x^2, y^2}$, and $B_{j_\text{TP}}^c$ denotes the set of true positive samples in the $j$-th bin for class $c$. Da-CE partitions predictions into $J$ bins and computes sample-level errors to avoid cancellation among directional mismatches. Since false positive samples lack deviation directions, Da-CE sets false positive errors to zero and uses only true positive samples for error computation.

\begin{table*}[t]
    \caption{\textbf{Comparison results with SOTA methods on COCO.} For evaluation, we train CR on COCO minival and evaluate CR on COCO minitest. \textbf{Bold} and \underline{underlined} values indicate the best and second-best results, respectively.}

    \renewcommand{\arraystretch}{1.35}
    
    \begin{adjustbox}{max width=\textwidth}
    \begin{tabular}{cl|cccc|cccc|cc}
    \specialrule{.1em}{.05em}{.05em}
    
    \multicolumn{2}{c|}{\multirow{2}{*}{\textrm{Method}}}
    & \multicolumn{4}{c|}{\textrm{Coordinate-wise}} & \multicolumn{4}{c|}{\textrm{Overall}} \\ 
    \cline{3-12}& 
    & \textrm{$\textrm{C}_{\textrm{x}_1}$-ECE$\downarrow$} & \textrm{$\textrm{C}_{\textrm{y}_1}$-ECE$\downarrow$} 
    & \textrm{$\textrm{C}_{\textrm{x}_2}$-ECE$\downarrow$} & \textrm{$\textrm{C}_{\textrm{y}_2}$-ECE$\downarrow$} 
    & $\textrm{D-ECE}\downarrow$ & \textrm{LaECE$\downarrow$} 
    & \textrm{LaECE$_0\downarrow$} & $\textrm{LaACE}_0\downarrow$ 
    & \textrm{AP$\uparrow$} & \textrm{LRP$\downarrow$}\\
    \specialrule{.1em}{.05em}{.05em}
    
    \multicolumn{2}{c|}{\textrm{Uncalibrated}~\footnotesize{\cite{zhu2020deformable}}}
    & 17.3 & 17.4 & 17.4 & 17.2
    & 14.9 & 12.2 & 12.7 & 27.1
    & 51.3 & 57.3 \\
    \specialrule{.1em}{.05em}{.05em}
    
    \multirow{3}{*}{\raisebox{1.5ex}{\rotatebox{270}{\textrm{{Train-time}}}}}
    & TCD~\footnotesize{\cite{munir2022towards}}
    & 18.0 & 18.1 & 18.1 & 17.5
    & 14.4 & 12.5 & 13.1 & 26.8
    & \underline{51.3} & \underline{57.1} \\
    & BPC~\footnotesize{\cite{munir2023bridging}}
    & 15.5 & 15.3 & 15.3 & 15.2
    & 11.3 & 12.3 & 12.8 & 25.4
    & 50.3 & 58.4 \\
    & Cal-DETR~\footnotesize{\cite{munir2023caldetr}}
    & 14.0 & 13.9 & 13.9 & 13.6
    & 9.8 & 11.7 & 11.7 & 24.6
    & \bf{52.5} & \bf{56.2} \\
    \specialrule{.1em}{.05em}{.05em}
    
    \multirow{2}{*}{\textrm{\rotatebox{270}{\shortstack{Post-hoc~~~~}}}}
    & IR for LaECE$_0$~\footnotesize{\cite{kuzucu2024calibration}}
    & 12.0 & 11.9 & 11.8 & 11.9
    & \underline{2.4} & \underline{8.4} & \bf{7.8} & \underline{23.1}
    & 51.0 & 57.3 \\
    & PS for LaECE$_0$~\footnotesize{\cite{kuzucu2024calibration}}
    & 13.9 & 14.0 & 14.0 & 13.8
    & \bf{2.3} & 10.0 & 9.7 & 23.5
    & \underline{51.3} & 57.3 \\

    &
    \cellcolor[HTML]{ECF4FF}\textrm{IR for Ours}
    & \cellcolor[HTML]{ECF4FF}\underline{7.7}
    & \cellcolor[HTML]{ECF4FF}\underline{10.5}
    & \cellcolor[HTML]{ECF4FF}\underline{11.4}
    & \cellcolor[HTML]{ECF4FF}\textbf{10.6}
    & \cellcolor[HTML]{ECF4FF}2.5
    & \cellcolor[HTML]{ECF4FF}\textbf{8.1}
    & \cellcolor[HTML]{ECF4FF}\underline{7.9}
    & \cellcolor[HTML]{ECF4FF}\textbf{22.7}
    & \cellcolor[HTML]{ECF4FF}50.4
    & \cellcolor[HTML]{ECF4FF}57.4
    \\

    &
    \cellcolor[HTML]{ECF4FF}\textrm{PS for Ours}
    & \cellcolor[HTML]{ECF4FF}\textbf{7.6}
    & \cellcolor[HTML]{ECF4FF}\textbf{10.4}
    & \cellcolor[HTML]{ECF4FF}\textbf{10.9}
    & \cellcolor[HTML]{ECF4FF}\underline{10.9}
    & \cellcolor[HTML]{ECF4FF}\underline{2.4}
    & \cellcolor[HTML]{ECF4FF}9.9
    & \cellcolor[HTML]{ECF4FF}10.0
    & \cellcolor[HTML]{ECF4FF}\underline{23.1}
    & \cellcolor[HTML]{ECF4FF}50.3
    & \cellcolor[HTML]{ECF4FF}57.3
    \\
    \specialrule{.1em}{.05em}{.05em}
    
    \end{tabular}
    \end{adjustbox}

    \label{tbl:main_com}
\end{table*}
\section{Experiments}
\label{sec:exp}
\subsubsection{Dataset}
We conduct experiments on COCO~\cite{lin2014microsoft} and Cityscapes~\cite{Cordts2016Cityscapes}. COCO contains 80 common object classes, while Cityscapes focuses on autonomous driving scenarios with 8 classes. To evaluate robustness under domain shift, we use COCO-C~\cite{hendrycks2019robustness}, which applies 3 types of corruptions to COCO, and Foggy Cityscapes~\cite{sakaridis2018semantic}, which simulates foggy conditions on Cityscapes. Further details are provided in the supplementary material.

\subsubsection{Evaluation Metric}

This work evaluates the calibration performance of fine-grained confidence scores using coordinate-wise expected calibration error (C-ECE) Eq.~\ref{eqn:c_ece} and direction-aware calibration error (Da-CE) Eq.~\ref{eqn:dauce}, and adopts established metrics from prior studies to assess the calibration performance of a calibrator on overall accuracy. The evaluation utilizes detection expected calibration error (D-ECE)~\cite{Kueppers_2020_CVPR_Workshops} ($\tau=0.5$), localization-aware expected calibration error (LaECE)~\cite{oksuz2023towards}($\tau=0.5$) and LaECE$_0$~\cite{kuzucu2024calibration}($\tau=0$). For evaluation, average precision (AP) is computed over the top-100 detections, and localization recall precision (LRP)~\cite{oksuz2018localization} is computed using the LRP threshold obtained on the validation set. Additional details are described in the supplementary material.

\subsubsection{Implementation Details}
For comparison with existing methods, we leverage deformable-DETR~\cite{zhu2020deformable} with ResNet-50~\cite{he2016deep}, which previous methods mainly utilized. 
The calibration procedure follows Kuzucu et al.~\cite{kuzucu2024calibration} and determines the calibration and operating thresholds by cross-validating LRP (IoU $\tau = 0$).

\subsection{Comparison with State-of-the-Art Methods}
\begin{table*}[t!]
\captionsetup{
  singlelinecheck=false
}
    \caption{\textbf{Comparison results with SOTA methods on Cityscapes dataset.} For the experiment, we fit the model on Cityscapes minival and evaluate the CR on Cityscapes minitest. 
    \textbf{Bold} and \underline{underlined} values indicate the best and second-best results, respectively.}

    \renewcommand{\arraystretch}{1.35}
    
    \begin{adjustbox}{max width=\textwidth}
    \begin{tabular}{cl|cccc|cccc|cc}
    \specialrule{.1em}{.05em}{.05em}
    
    \multicolumn{2}{c|}{\multirow{2}{*}{\textrm{Method}}}
    & \multicolumn{4}{c|}{\textrm{Coordinate-wise}} & \multicolumn{4}{c|}{\textrm{Overall}} \\ 
    \cline{3-12}& 
    & \textrm{$\textrm{C}_{\textrm{x}_1}$-ECE$\downarrow$} & \textrm{$\textrm{C}_{\textrm{y}_1}$-ECE$\downarrow$} 
    & \textrm{$\textrm{C}_{\textrm{x}_2}$-ECE$\downarrow$} & \textrm{$\textrm{C}_{\textrm{y}_2}$-ECE$\downarrow$} 
    & $\textrm{D-ECE}\downarrow$ & \textrm{LaECE$\downarrow$}
    & \textrm{LaECE$_0\downarrow$} & $\textrm{LaACE}_0\downarrow$ 
    & \textrm{AP$\uparrow$} & \textrm{LRP$\downarrow$}\\
    \specialrule{.1em}{.05em}{.05em}
    
    \multicolumn{2}{c|}{\textrm{Uncalibrated}~\footnotesize{\cite{zhu2020deformable}}}
    & 13.7 & 15.5 & 14.2 & 15.5
    & 13.7 & 11.8 & 13.4 & 30.8
    & 44.5 & 66.4 \\
    \specialrule{.1em}{.05em}{.05em}
    
    \multirow{3}{*}{\raisebox{1.5ex}{\rotatebox{270}{\textrm{{Train-time}}}}}
    & TCD~\footnotesize{\cite{munir2022towards}}
    & 12.5 & 13.7 & 12.4 & 12.8
    & 15.3 & 11.7 & 12.6 & 29.3
    & 29.1 & 76.9 \\
    & BPC~\footnotesize{\cite{munir2023bridging}}
    & 13.8 & 15.4 & 14.0 & 15.8
    & 5.4 & 13.7 & 14.2 & \bf{26.6}
    & 30.5 & 74.2 \\
    & Cal-DETR~\footnotesize{\cite{munir2023caldetr}}
    & 13.4 & 15.1 & 13.4 & 14.5
    & 13.0 & 11.0 & 12.7 & 29.4
    & 38.7 & 70.6 \\
    \specialrule{.1em}{.05em}{.05em}
    
    \multirow{4}{*}{\textrm{\rotatebox{270}{\shortstack{Post-hoc~~~~}}}}
    & IR for LaECE$_0$~\footnotesize{\cite{kuzucu2024calibration}}
    & 12.2 & 13.6 & 11.6 & 13.9
    & 1.5 & \underline{8.0} & \underline{7.5} & \underline{27.2}
    & \underline{43.5} & \bf{66.4} \\
    & PS for LaECE$_0$~\footnotesize{\cite{kuzucu2024calibration}}
    & 12.9 & 14.7 & 13.1 & 15.1
    & \bf{1.0} & 10.0 & 9.6 & 27.4
    & \bf{44.5} & \bf{66.4} \\
    
    &
    \cellcolor[HTML]{ECF4FF}\textrm{IR for Ours}
    & \cellcolor[HTML]{ECF4FF}\underline{6.9}
    & \cellcolor[HTML]{ECF4FF}\textbf{11.1}
    & \cellcolor[HTML]{ECF4FF}\textbf{10.3}
    & \cellcolor[HTML]{ECF4FF}\underline{11.3}
    & \cellcolor[HTML]{ECF4FF}1.2
    & \cellcolor[HTML]{ECF4FF}\textbf{6.2}
    & \cellcolor[HTML]{ECF4FF}\textbf{6.5}
    & \cellcolor[HTML]{ECF4FF}29.7
    & \cellcolor[HTML]{ECF4FF}42.5
    & \cellcolor[HTML]{ECF4FF}67.0
    \\
    
    &
    \cellcolor[HTML]{ECF4FF}\textrm{PS for Ours}
    & \cellcolor[HTML]{ECF4FF}\textbf{6.5}
    & \cellcolor[HTML]{ECF4FF}\underline{11.4}
    & \cellcolor[HTML]{ECF4FF}\underline{10.5}
    & \cellcolor[HTML]{ECF4FF}\textbf{10.6}
    & \cellcolor[HTML]{ECF4FF}\underline{1.1}
    & \cellcolor[HTML]{ECF4FF}9.6
    & \cellcolor[HTML]{ECF4FF}9.4
    & \cellcolor[HTML]{ECF4FF}27.6
    & \cellcolor[HTML]{ECF4FF}42.0
    & \cellcolor[HTML]{ECF4FF}\underline{66.9}
    \\
    \specialrule{.1em}{.05em}{.05em}
    
    \end{tabular}
    \end{adjustbox}

    \label{tbl:city_com}
\end{table*}

\begin{table*}[t!]
    \caption{
        \textbf{Comparison with SOTA methods on corrupted COCO dataset for domain shift scenarios.} To simulate a domain shift scenario, we train the confidence re-encoder on COCO minival and report inference results on COCO-C minitest. \textbf{Bold} and \underline{underlined} values indicate the best and second-best results, respectively.
    }
    \renewcommand{\arraystretch}{1.35}
    
    \begin{adjustbox}{max width=\textwidth}
    \begin{tabular}{cl|cccc|cccc|cc}
    \specialrule{.1em}{.05em}{.05em}
    
    \multicolumn{2}{c|}{\multirow{2}{*}{\textrm{Method}}}
    & \multicolumn{4}{c|}{\textrm{Coordinate-wise}} & \multicolumn{4}{c|}{\textrm{Overall}} \\ 
    \cline{3-12}& 
    & \textrm{$\textrm{C}_{\textrm{x}_1}$-ECE$\downarrow$} & \textrm{$\textrm{C}_{\textrm{y}_1}$-ECE$\downarrow$} 
    & \textrm{$\textrm{C}_{\textrm{x}_2}$-ECE$\downarrow$} & \textrm{$\textrm{C}_{\textrm{y}_2}$-ECE$\downarrow$} 
    & $\textrm{D-ECE}\downarrow$ & \textrm{LaECE$\downarrow$} 
    & \textrm{LaECE$_0\downarrow$} & $\textrm{LaACE}_0\downarrow$ 
    & \textrm{AP$\uparrow$} & \textrm{LRP$\downarrow$}\\
    \specialrule{.1em}{.05em}{.05em}
    
    \multicolumn{2}{c|}{\textrm{Uncalibrated}~\footnotesize{\cite{zhu2020deformable}}}
    & 20.3 & 20.3 & 20.2 & 20.2
    & 15.9 & 15.2 & 15.2 & 29.4
    & 30.5 & 74.0 \\
    \specialrule{.1em}{.05em}{.05em}
    
    \multirow{3}{*}{\raisebox{1.5ex}{\rotatebox{270}{\textrm{{Train-time}}}}}
    & TCD~\footnotesize{\cite{munir2022towards}}
    & 20.6 & 20.6 & 20.4 & 20.3
    & 16.4 & 14.9 & 14.8 & 28.9
    & 29.7 & 74.4 \\
    & BPC~\footnotesize{\cite{munir2023bridging}}
    & 19.0 & 19.0 & 19.0 & 18.9
    & 13.0 & 15.0 & 14.5 & 27.5
    & 29.6 & 75.1 \\
    & Cal-DETR~\footnotesize{\cite{munir2023caldetr}}
    & 17.8 & 18.0 & 17.7 & 17.7
    & 11.3 & 15.4 & 14.6 & 27.0
    & \bf{30.5} & \bf{73.9} \\
    \specialrule{.1em}{.05em}{.05em}
    
    \multirow{4}{*}{\textrm{\rotatebox{270}{\shortstack{Post-hoc~~~~}}}}
    & IR for LaECE$_0$~\footnotesize{\cite{kuzucu2024calibration}}
    & 16.1 & 16.3 & 16.0 & 16.1
    & 3.1 & \underline{11.7} & \underline{11.2} & \underline{26.2}
    & \underline{30.2} & \underline{74.0} \\
    & PS for LaECE$_0$~\footnotesize{\cite{kuzucu2024calibration}}
    & 18.5 & 18.6 & 18.4 & 18.5
    & \bf{2.6} & 14.4 & 13.7 & 26.4
    & \bf{30.5} & \underline{74.0} \\
    
    &
    \cellcolor[HTML]{ECF4FF}\textrm{IR for Ours}
    & \cellcolor[HTML]{ECF4FF}\underline{11.9}
    & \cellcolor[HTML]{ECF4FF}\underline{15.1}
    & \cellcolor[HTML]{ECF4FF}\underline{15.4}
    & \cellcolor[HTML]{ECF4FF}\textbf{13.9}
    & \cellcolor[HTML]{ECF4FF}\underline{2.9}
    & \cellcolor[HTML]{ECF4FF}\textbf{11.5}
    & \cellcolor[HTML]{ECF4FF}\textbf{11.0}
    & \cellcolor[HTML]{ECF4FF}\underline{26.2}
    & \cellcolor[HTML]{ECF4FF}29.8
    & \cellcolor[HTML]{ECF4FF}74.1
    \\
    
    &
    \cellcolor[HTML]{ECF4FF}\textrm{PS for Ours}
    & \cellcolor[HTML]{ECF4FF}\textbf{11.7}
    & \cellcolor[HTML]{ECF4FF}\textbf{15.0}
    & \cellcolor[HTML]{ECF4FF}\textbf{15.3}
    & \cellcolor[HTML]{ECF4FF}\underline{14.4}
    & \cellcolor[HTML]{ECF4FF}\textbf{2.6}
    & \cellcolor[HTML]{ECF4FF}15.3
    & \cellcolor[HTML]{ECF4FF}14.1
    & \cellcolor[HTML]{ECF4FF}\textbf{26.1}
    & \cellcolor[HTML]{ECF4FF}29.6
    & \cellcolor[HTML]{ECF4FF}74.3
    \\
    \specialrule{.1em}{.05em}{.05em}
    
    \end{tabular}
    \end{adjustbox}

    \label{tbl:domain_com}
\end{table*}

\begin{table*}[t!]
    \caption{
        \textbf{Comparison results with SOTA methods on Foggy Cityscapes.} To simulate a domain shift scenario, we train the confidence re-encoder on Cityscapes minival and report the inference results on Foggy Cityscapes minitest. \textbf{Bold} and \underline{underlined} values indicate the best and second-best results, respectively.
    }
    \renewcommand{\arraystretch}{1.35}
    
    \begin{adjustbox}{max width=\textwidth}
    \begin{tabular}{cl|cccc|cccc|cc}
    \specialrule{.1em}{.05em}{.05em}
    
    \multicolumn{2}{c|}{\multirow{2}{*}{\textrm{Method}}}
    & \multicolumn{4}{c|}{\textrm{Coordinate-wise}} & \multicolumn{4}{c|}{\textrm{Overall}} \\ 
    \cline{3-12}& 
    & \textrm{$\textrm{C}_{\textrm{x}_1}$-ECE$\downarrow$} & \textrm{$\textrm{C}_{\textrm{y}_1}$-ECE$\downarrow$} 
    & \textrm{$\textrm{C}_{\textrm{x}_2}$-ECE$\downarrow$} & \textrm{$\textrm{C}_{\textrm{y}_2}$-ECE$\downarrow$} 
    & $\textrm{D-ECE}\downarrow$ & \textrm{LaECE$\downarrow$} 
    & \textrm{LaECE$_0\downarrow$} & $\textrm{LaACE}_0\downarrow$ 
    & \textrm{AP$\uparrow$} & \textrm{LRP$\downarrow$}\\
    \specialrule{.1em}{.05em}{.05em}
    
    \multicolumn{2}{c|}{\textrm{Uncalibrated}~\footnotesize{\cite{zhu2020deformable}}}
    & 16.7 & 18.6 & 16.9 & 18.4
    & 13.9 & 13.5 & 13.8 & 28.8
    & 31.0 & 74.7 \\
    \specialrule{.1em}{.05em}{.05em}
    
    \multirow{3}{*}{\raisebox{1.5ex}{\rotatebox{270}{\textrm{{Train-time}}}}}
    & TCD~\footnotesize{\cite{munir2022towards}}
    & 16.7 & 18.2 & 16.7 & 18.0
    & 18.0 & 12.8 & 13.4 & 28.3
    & 23.6 & 81.1 \\
    & BPC~\footnotesize{\cite{munir2023bridging}}
    & 16.9 & 18.0 & 17.0 & 18.8
    & 5.2 & 17.5 & 17.2 & 28.7
    & 23.7 & 79.9 \\
    & Cal-DETR~\footnotesize{\cite{munir2023caldetr}}
    & 16.8 & 18.6 & 16.8 & 18.3
    & 12.4 & 12.8 & 13.8 & 28.4
    & 28.4 & 76.9 \\
    \specialrule{.1em}{.05em}{.05em}
    
    \multirow{4}{*}{\textrm{\rotatebox{270}{\shortstack{Post-hoc~~~~}}}}
    & IR for LaECE$_0$~\footnotesize{\cite{kuzucu2024calibration}}
    & \underline{13.7} & 15.8 & 14.1 & 15.8
    & 1.3 & \bf{7.8} & \bf{7.5} & \underline{23.9}
    & \underline{30.9} & \bf{74.6} \\
    & PS for LaECE$_0$~\footnotesize{\cite{kuzucu2024calibration}}
    & 17.1 & 19.4 & 17.3 & 19.0
    & \bf{1.0} & 12.0 & 11.6 & 24.4
    & \bf{31.0} & \underline{74.7} \\
    
    &
    \cellcolor[HTML]{ECF4FF}\textrm{IR for Ours}
    & \cellcolor[HTML]{ECF4FF}\textbf{10.2}
    & \cellcolor[HTML]{ECF4FF}\underline{13.5}
    & \cellcolor[HTML]{ECF4FF}\underline{12.9}
    & \cellcolor[HTML]{ECF4FF}\underline{11.7}
    & \cellcolor[HTML]{ECF4FF}\underline{1.2}
    & \cellcolor[HTML]{ECF4FF}\underline{8.7}
    & \cellcolor[HTML]{ECF4FF}\underline{8.4}
    & \cellcolor[HTML]{ECF4FF}\textbf{22.2}
    & \cellcolor[HTML]{ECF4FF}30.3
    & \cellcolor[HTML]{ECF4FF}74.8
    \\
    
    &
    \cellcolor[HTML]{ECF4FF}\textrm{PS for Ours}
    & \cellcolor[HTML]{ECF4FF}\textbf{10.2}
    & \cellcolor[HTML]{ECF4FF}\textbf{13.3}
    & \cellcolor[HTML]{ECF4FF}\textbf{12.7}
    & \cellcolor[HTML]{ECF4FF}\textbf{11.6}
    & \cellcolor[HTML]{ECF4FF}1.4
    & \cellcolor[HTML]{ECF4FF}11.8
    & \cellcolor[HTML]{ECF4FF}10.6
    & \cellcolor[HTML]{ECF4FF}25.0
    & \cellcolor[HTML]{ECF4FF}30.8
    & \cellcolor[HTML]{ECF4FF}74.9
    \\
    \specialrule{.1em}{.05em}{.05em}
    \end{tabular}
    \end{adjustbox}

    \label{tbl:city_domain}
\end{table*}

\subsubsection{Common Object Scenarios}
Tab.~\ref{tbl:main_com} presents comparison results with SOTA methods on COCO. While prior approaches achieve impressive box-level calibration performance in terms of D-ECE, LaECE, and LaECE$_0$, a single calibrated confidence score fails to capture the diverse distributions of CARs across bounding box coordinates, resulting in limited alignment with coordinate-wise localization accuracy. In contrast, the coordinate-wise confidence scores estimated by CR utilizing PS achieves an average C-ECE of 10.0 over prior methods, while the box-level calibration performance obtained from the approximated IoU remains competitive with existing approaches.

\subsubsection{Autonomous Driving Scenarios}
Tab.~\ref{tbl:city_com} presents comparison results with SOTA methods on Cityscapes. Prior approaches achieve well-calibrated box-level performance, but coordinate-wise calibration remains inadequate. For example, PS for LaECE$_0$~\cite{kuzucu2024calibration} records 9.6 in LaECE$_0$, yet shows 14.7 in C${y^1}$-ECE and 15.1 in C${y^2}$-ECE. This result indicates poor alignment with coordinate-wise localization accuracy. In contrast, CR with PS achieves an average C-ECE of 9.8 across all coordinates, demonstrating consistent coordinate-wise calibration.

\begin{table}[t!]
    \caption{\textbf{Ablation results of direction influence for C-ECE} For the ablation study, we compare results on the COCO dataset over five random seeds with and without prediction deviation direction estimation. \textbf{Bold} indicates the best results.}
    \label{abl_direc}
        \resizebox{\linewidth}{!}{
        \begin{tabular}{c | cccc | cc} 
            \specialrule{.1em}{.05em}{.05em}
            \textbf{Direction} &  
            $\bar{\textrm{C}_{\textrm{x}_1}}$-ECE$\downarrow$ & 
            $\bar{\textrm{C}_{\textrm{y}_1}}$-ECE$\downarrow$ & 
            $\bar{\textrm{C}_{\textrm{x}_2}}$-ECE$\downarrow$ & 
            $\bar{\textrm{C}_{\textrm{y}_2}}$-ECE$\downarrow$ & 
            $\bar{\textrm{LaECE$_0\downarrow$}}$ & $\bar{\textrm{LaACE$_0\downarrow$}}$ \\ 
            \specialrule{.1em}{.05em}{.05em}
            - & 11.22 ($\pm$ 0.058) & 11.90 ($\pm$ 0.106) & 10.42 ($\pm$ 0.054) & 10.62 ($\pm$ 0.134) & 9.96 ($\pm$ 0.022) & \textbf{23.14} ($\pm$ 0.034) \\
            \ding{51} & \textbf{11.18} ($\pm$ 0.038) & \textbf{10.86} ($\pm$ 0.042) & \textbf{10.26} ($\pm$ 0.074) & \textbf{10.58} ($\pm$ 0.106) & \textbf{9.88} ($\pm$ 0.022) & \textbf{23.14} ($\pm$ 0.002) \\
            \specialrule{.1em}{.05em}{.05em}
        \end{tabular}}
\end{table}

\subsubsection{Domain Shift}
Tab.~\ref{tbl:domain_com} and Tab.~\ref{tbl:city_domain} present results under 
domain shift on COCO-C and Foggy Cityscapes, respectively. While existing methods 
maintain competitive box-level calibration performance under domain shift, they 
exhibit degraded coordinate-wise calibration, revealing limited ability to 
represent per-coordinate localization accuracy in unseen domains. In contrast, CR consistently achieves superior coordinate-wise calibration, recording an average C-ECE of 14.1 on COCO-C and 12.0 on Foggy Cityscapes, while maintaining competitive box-level calibration performance using PS, comparable to existing methods.

\subsection{Analysis of Rethinking Detection Calibration}
\subsubsection{Ablation Study}
Tab.~\ref{abl_direc} presents the ablation results over five random seeds with and without directional displacement estimation (DDE). The model without DDE does not predict deviation directions, and the IoU approximation therefore excludes directional information. The lower LaECE$_0$ and LaACE$_0$ values show that directional information improves box-level IoU approximation and calibration. The improved box-level calibration also enables the estimation of a more appropriate operating threshold for predictions. As a result, DDE improves coordinate-level calibration in terms of C-ECE and reduces variance across random seeds, indicating more robust prediction.

\begin{figure*}[t!]
    \centering
    \includegraphics[width=1\linewidth]{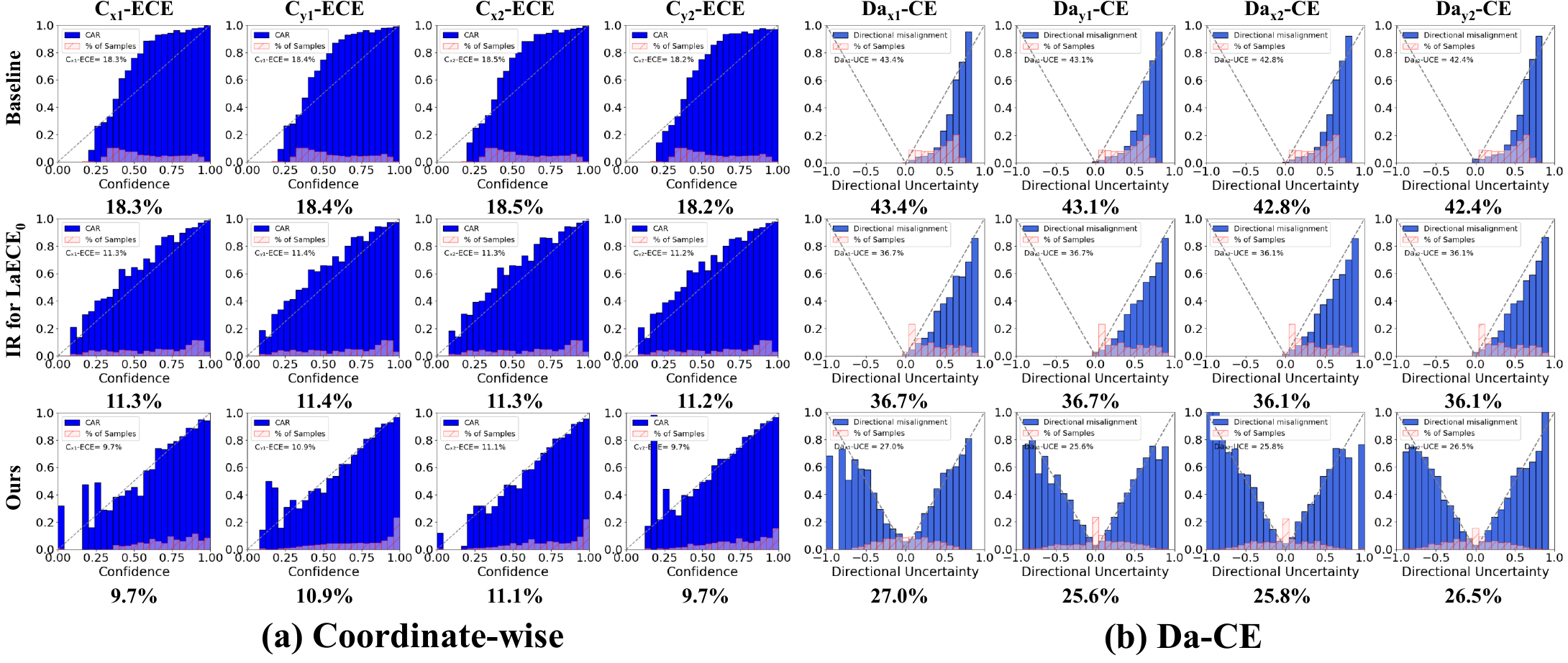}
    \caption{
        \textbf{Reliability Diagram}. (a) C-ECE diagrams plot average CAR against predicted confidence per bin. (b) Da-CE diagrams plot average misalignment against predicted directional mismatch. Perfect calibration corresponds to alignment with the diagonal.
    }
    \label{fig:RD}
\end{figure*}

\subsubsection{Comparison with GP-Normal}
GP-Normal~\cite{Kueppers_2022_ECCV_Workshops} performs calibration using probabilistic object detectors that estimate uncertainty through coordinate-wise probability distributions and align the estimated uncertainty with actual localization errors. 
For a fair comparison with GP-Normal, which estimates uncertainty via coordinate-wise probability distributions, we measure correlation with coordinate-wise error, using normalized inverse standard deviation for GP-Normal and calibrated confidence scores for ReDC. ReDC achieves a higher mean correlation than GP-Normal (0.1771 vs. 0.0263), indicating that ReDC better captures coordinate-wise errors.

\begin{wrapfigure}{t!}{0.60\linewidth}
    \centering
    \includegraphics[width=1\linewidth]{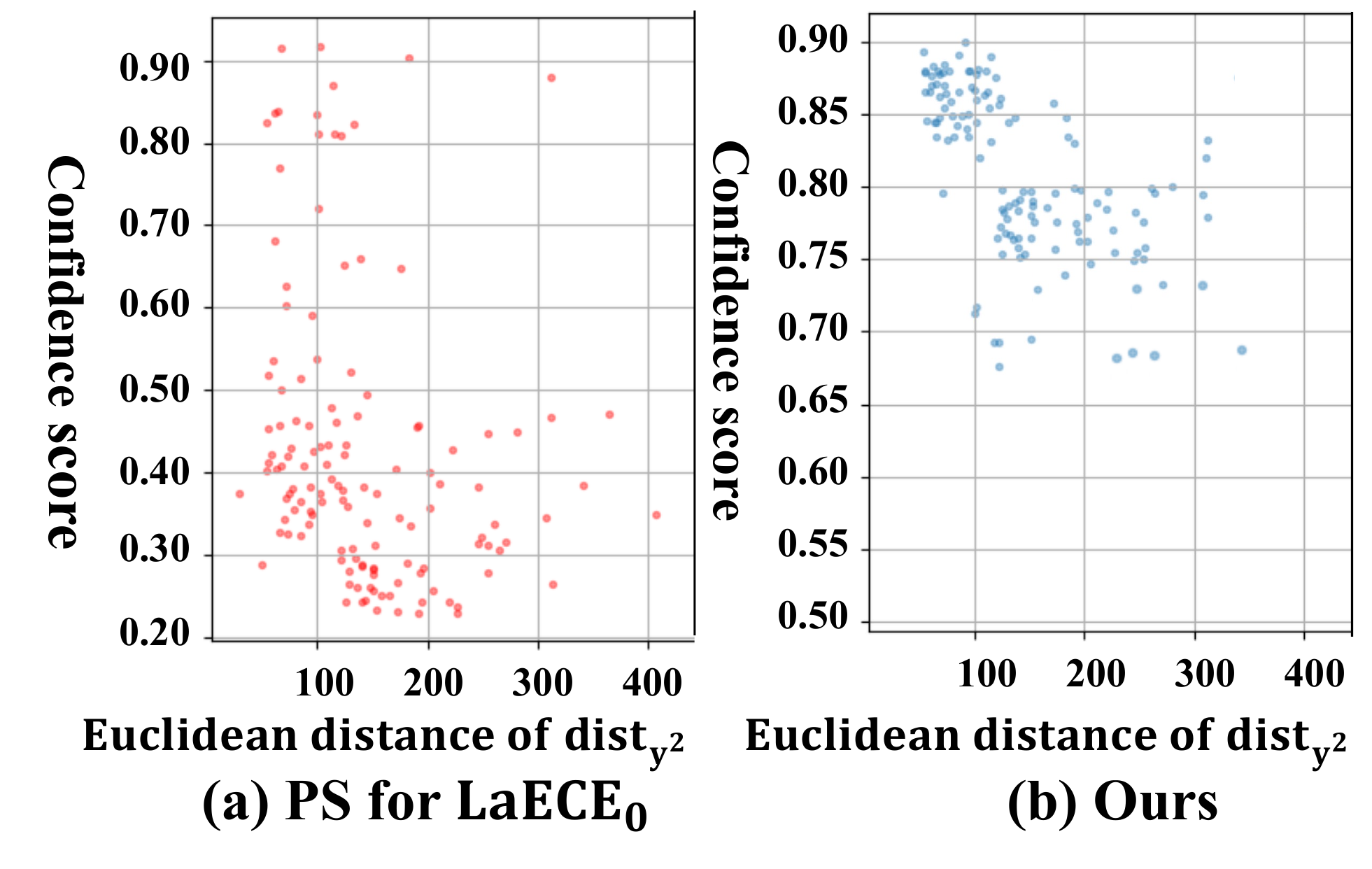}
    \caption{\textbf{Analysis of $y^2$ confidence scores for samples exhibiting large errors in the $y^2$ coordinate.} Samples selected from the top 3\% per class exhibiting large errors between ground-truth and predicted bounding boxes along the $y^2$ coordinate.}\label{fig:scatter}
\end{wrapfigure}

\subsubsection{Correlation Analysis of $y^2$ Coordinate Error}
Fig.~\ref{fig:scatter} shows coordinate-wise confidence distributions for boxes with large localization errors along $y^2$. As shown in Fig.~\ref{fig:scatter}-As shown in (a), prior methods rely on a single confidence score and fail to reflect increasing localization error. In contrast, (b) shows ReDC's confidence score negatively correlates with pixel distance error along $y^2$, indicating that ReDC quantitatively captures positional confidence and conveys positional information rarely provided by conventional 2D detection frameworks.

\subsubsection{Reliable Diagram}
In Fig.~\ref{fig:RD}, reliability diagrams compare ReDC with IR for LaECE₀ and the baseline (DINO). Since the baseline and IR for LaECE₀ only capture box-level accuracy, they produce under-confident coordinate-wise predictions, as actual coordinate-wise accuracy exceeds the empirical IoU distribution. ReDC, in contrast, achieves well-calibrated coordinate-wise confidence while maintaining competitive box-level calibration. The Da-CE diagrams further reveal systematic directional bias in the baseline and IR for LaECE₀, whose predicted mismatch directions fail to align with actual coordinate-wise error directions. ReDC substantially reduces across all coordinates.

\begin{table}[t]
    \centering
    \caption{\textbf{Model-agnostic post-hoc confidence calibration.} \textbf{Bold} and \underline{underlined} values indicate the best and second-best results, respectively.}
    
    \scriptsize
    \setlength{\tabcolsep}{3pt}
    
    \begin{tabular}{clc|cccc|c}
    \specialrule{.1em}{.05em}{.05em}
    & Method & Type
    & \textrm{$\textrm{C}_{\textrm{x}_1}$-ECE$\downarrow$} & \textrm{$\textrm{C}_{\textrm{y}_1}$-ECE$\downarrow$} 
    & \textrm{$\textrm{C}_{\textrm{x}_2}$-ECE$\downarrow$} & \textrm{$\textrm{C}_{\textrm{y}_2}$-ECE$\downarrow$}
    & \textrm{Da-CE$\downarrow$}\\
    \specialrule{.1em}{.05em}{.05em}
    
    \multirow{3}{*}{\rotatebox{270}{Base~~~}}
    & VFNET~\cite{zhang2021varifocalnet} & 1 & 22.9 & 22.9 & 22.7 & 22.4 & 48.1 \\
    & Cascade R-CNN~\cite{cai18cascadercnn} & 2 & 16.5 & 16.0 & 16.3 & 16.4 & \bf{19.3} \\
    & DINO~\cite{zhang2023dino} & ViT & 18.3 & 18.4 & 18.5 & 18.2 & 42.9 \\
    
    \specialrule{.1em}{.05em}{.05em}
    
    \multirow{3}{*}{\rotatebox{270}{Ours~~~}}
    & VFNET~\cite{zhang2021varifocalnet} & 1 & \underline{12.9} & 12.9 & 14.2 & 13.0 & \underline{25.7} \\
    & Cascade R-CNN~\cite{cai18cascadercnn} & 2 & \underline{10.6} & \bf{11.0} & \underline{12.0} & \underline{11.8} & 28.1 \\
    & DINO~\cite{zhang2023dino} & ViT & \bf{9.4} & \bf{11.0} & \bf{11.1} & \bf{9.6} & 26.0 \\
    \specialrule{.1em}{.05em}{.05em}
    
    \end{tabular}
    \label{tbl:sev_det}
\end{table}

\subsubsection{Model-agnostic Coordinate-wise Calibration}
Tab.~\ref{tbl:sev_det} reports coordinate-wise calibration results across diverse detector architectures, including one-stage, two-stage, and transformer-based detectors. Since CR leverages bounding box features and logits that any detector already produces, it requires no architectural modifications and achieves consistent coordinate-wise calibration performance regardless of the detector.

\section{Conclusion}
\label{sec:con}
We propose ReDC, a post-hoc calibration framework that provides coordinate-wise confidence scores for object detection. We introduce CAR to measure localization accuracy at each coordinate and train a lightweight confidence re-encoder to align coordinate-wise confidence scores with CAR. ReDC further captures the directional characteristics of localization errors through DDE, and we introduce C-ECE and Da-CE as new metrics that jointly assess coordinate-level and directional accuracy. Extensive experiments on COCO and Cityscapes, under both in-domain and domain shift scenarios, show that ReDC outperforms existing methods in coordinate-wise calibration while achieving competitive box-level performance across diverse detector architectures.

\section*{Acknowledgements}
This work was partly supported by the Institute of Information \& Communications Technology Planning \& Evaluation (IITP) grant funded by the Korea government (MSIT) [RS-2021-II211341, Artificial Intelligence Graduate School Program (Chung-Ang University); No.RS-2024-00437576, Development of autonomous performance improvement technology for video surveillance system based on edge-analysis server connection] and a grant (22193MFDS471) from the Ministry of Food and Drug Safety in 2024.

\bibliographystyle{splncs04}
\bibliography{main}

\end{document}